# Open Source Computer Vision-based Layer-wise 3D Printing Analysis

Aliaksei L. Petsiuk[1] and Joshua M. Pearce[1,2,3]

[1]Department of Electrical & Computer Engineering, Michigan Technological University, Houghton, MI 49931, USA
[2]Department of Material Science & Engineering, Michigan Technological University, Houghton, MI 49931, USA
[3]Department of Electronics and Nanoengineering, School of Electrical Engineering, Aalto University, Espoo, FI-00076, Finland

apetsiuk@mtu.edu, pearce@mtu.edu

**Graphical Abstract**

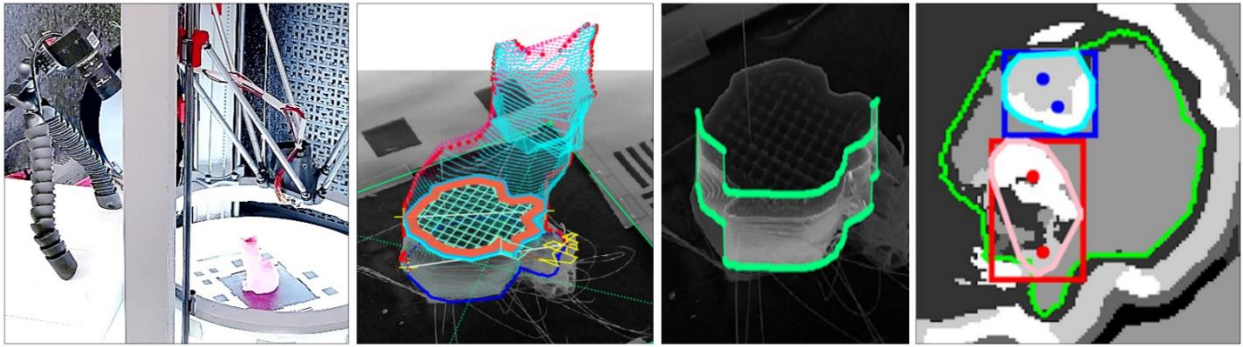

**Highlights**
- Developed a visual servoing platform using a monocular multistage image segmentation
- Presented algorithm prevents critical failures during additive manufacturing
- The developed system allows tracking printing errors on the interior and exterior

## Abstract
The paper describes an open source computer vision-based hardware structure and software algorithm, which analyzes layer-wise the 3-D printing processes, tracks printing errors, and generates appropriate printer actions to improve reliability. This approach is built upon multiple-stage monocular image examination, which allows monitoring both the external shape of the printed object and internal structure of its layers. Starting with the side-view height validation, the developed program analyzes the virtual top view for outer shell contour correspondence using the multi-template matching and iterative closest point algorithms, as well as inner layer texture quality clustering the spatial-frequency filter responses with Gaussian mixture models and segmenting structural anomalies with the agglomerative hierarchical clustering algorithm. This allows evaluation of both global and local parameters of the printing modes. The experimentally-verified analysis time per layer is less than one minute, which can be considered a quasi-real-time process for large prints. The systems can work as an intelligent printing suspension tool designed to save time and material. However, the results show the algorithm provides a means to systematize in situ printing data as a first step in a fully open source failure correction algorithm for additive manufacturing.

## 1. Introduction

Despite a long evolution of additive manufacturing (AM), starting from the first patent in 1971 [1], 3-D printing technology has only recently exploded in popularity due to the radical decreases in costs brought on by the introduction of the self-replicating rapid prototyper (RepRap) 3-D printer [2-4]. With the generalized material extrusion printing process called fused filament fabrication (FFF) technology gaining prominence in the field with the expiration of fused deposition modeling (FDM) patents, FFF now dominates the 3-D printing market for printers in use [5]. Making AM accessible to the masses of consumers has enabled the emergence of a distributed manufacturing paradigm [6-14], where 3-D printing can be used to manufacture open source products for the consumer and by the consumer directly for less (in many cases more than an order of magnitude less) money than purchasing of mass-manufactured proprietary products [10, 15-18]. The downloaded substitution values [19, 20] for digital manufacturing with AM of even sophisticated high-end products [21-24] provides a high return on investment [25]. In addition, there is some evidence that AM distributed manufacturing reduces the impact on the environment [26-30]. However, both the economics and environmental impact of distributed manufacturing is heavily impacted by success rate. Early work on self-built RepRaps estimated a 20% failure rate [9] and more recent values of about 10% [31], which is in the range (1-20%) of recent polling of the Reddit r/3Dprinting community [32].

Although the cost of printing with an FFF-based 3-D printer is trivial compared to the other AM techniques, printing errors are substantial enough to impact the economic and environmental merits of the approach. To this end, several studies and techniques have been attempted to reduce failure rates. Nuchitprasitchai et al. [33] were able to detect the "blocked nozzle" and "incomplete print" failures for six different objects in five colors. The printed objects have been tested in a single- and double-camera experiments to determine critical 5% deviations in shape and size at every 50$^{th}$ layer. In the subsequent work [34], the authors used three pairs of cameras 120 degrees apart to reconstruct the 3-D surface points of the printed objects at every 30th layer to detect critical 10% size deviations. Garanger et al. [35] implemented a closed-loop control system for a certain additive manufacturing process to reach the acceptable stiffness for leaf spring-like objects. Delli and Chang [36] proposed a binary (failed/not failed) 3-D printing error classification approach based on a supervised machine learning technique, where the quality check is being performed at critical stages during the printing process. Fastowicz and Okarma [37] developed a quality assessment method based on texture analysis using the Gray-Level Co-occurrence Matrix [38-40] and selected Haralick features [41]. In [42], Cummings et al., developed a closed-loop control framework that detects and corrects filament bonding failures by using the ultrasonic sensor and manipulating the print bed temperature during the printing process. Finally, Rao et al. [43] developed a heterogeneous sensor framework for real-time surface roughness analysis based on such printing parameters as extruder temperature, layer thickness on build quality, and feed to flow ratio.

In the more mature areas of AM with higher-priced materials and 3-D printers, various methods of quality control have been instituted to minimize print failure. Scime and Beth [44] introduced an in-situ anomaly detection approach based on the unsupervised machine learning technique for laser powder bed fusion (LPBF) additive manufacturing. The developed method [44] determines the possible causes for partial fusion failures and provides potential strategies to build quality enhancement in the future. Xiong et al. [45] developed a camera-based system in gas metal arc welding to monitor and control the distance between the nozzle and the top surface by compensating such parameters as the deposition rate and working flat level. Nassar et al. [46] developed a temperature-based inner layer control strategy and analyzed its effects on the hardness and microstructure of the metal printed component. In both [45] and [46], the authors developed closed-loop control capabilities, but the methods, however, focus on microscopic properties without considering the global macrostructure of the object being built. Okaro et al. [47] introduced a semi-supervised machine learning algorithm based on large sets of photodiode data for automatic "faulty" and "acceptable" tensile strength assessment in laser power bed fusion additive manufacturing. Garanger et al. [48] suggested a number of semantic rules within 3-D printing files, which provide desired specifications and, based on material properties, real-time topology and finite element analysis, generate feedback laws for the control system. Yuan et al. [49] developed a two-step machine learning approach to real-time laser track welds assessment in LPBF processes. Sitthi-Amorn et al. [50] introduced a layer-wise correction technique for multi-material 3-D printing. Razaviarab et al. [51] proposed a closed-loop machine learning algorithm to detect layer defects in metal printing.

The key point of the previous works is addressing a limited number of specific cases of local defects without taking into account the global parameters of the printed parts (such as full-scale shift or deformation, deviation of the dimensions of the parts from the calculated ones, etc.). Most of the methods described also do not imply an on-the-fly algorithm for compensating, correcting or eliminating manufacturing failures.

In order to mature the FFF-based 3-D printing quality control to reach that of the more expensive AM technologies, this study presents and tests a free and open source software algorithm and hardware structure based on computer vision, which allows layer-wise analysis of 3-D printing processes to segment and track manufacturing errors, and repairing procedures to generate appropriate printer actions during fabrication. These repair-based actions are designed to minimize the waste of printing material and printing time caused by erroneous printing attempts. The approach is based on monocular multiple-stage image processing that monitors the external shape of the printed object and internal structure of the layers. The developed framework analyzes both global (deformation of overall dimensions) and local (deformation of filling) deviations of print modes, it restores the level of scale and displacement of the deformed layer and introduces a potential opportunity of repairing internal defects in printed layers. The analysis time as a function of layers is quantified and the results are discussed.

## 2. Visual Servoing Platform Design

FFF/FDM 3-D printing is the simplest and most common form of additive manufacturing [52]. Increasing the resiliency and quality of FFF printing will thus provide a significant benefit to makers' communities around the world.

Table 1 summarizes FFF 3-D printing errors and methods for their elimination guided by makers' experience with low-cost desktop 3-D printers [53, 54]. As can be seen from Table 1, print options that affect the possibility of an error can be divided into three categories:
1. mechanical parameters of the 3-D printer and its environment,
2. temperature parameters of the fan/nozzle and the printing bed,
3. algorithm for converting a Standard Tessellation Language (STL) model to G-Code instructions (i.e. slicer parameters).

The temperature conditions, feed rate, traveling speed of the extruder, as well as some parameters of the slicing algorithm (such as height of the printed layer, thickness of lines and percentage of their overlapping, etc.) can be controlled by the G-Code commands. Thus, by adapting the G-Code parameters during printing, it is possible to indirectly compensate for the shortcomings of the temperature regime and slicer parameters. The mechanical parameters of the printer (stability of assembly, the presence of grease in moving parts, belt tension, the electrical voltage of stepper motor drivers, etc.), as well as flaws of the CAD model design, are practically impossible to compensate for during printing. However, having the ability to vary certain 3-D printing options, a computer algorithm can be written to eliminate or reduce the likelihood of failures from the former root causes, the latter need to be addressed by physically fixing the machine.

The existing market for FFF 3-D printers is very diverse and is represented by models of various shapes with both moving and stationary working surfaces (printing beds). To test the algorithm, a delta-type RepRap printer was chosen [55], which is an open-source development with a fixed working surface, which simplifies the synchronization of the camera with printing processes.

A Michigan Tech Open Sustainability Technology (MOST) Delta RepRap FFF-based 3-D printer [55] with a 250 mm diameter and 240 mm high cylindrical working volume was used (Figure 1). It fuses 1.75 mm polylactic acid (PLA) plastic filament under a temperature of 200 °C from a nozzle with a 0.4 mm diameter. The printer operates by RAMPS 1.4 printer controller with an integrated Secure Digital (SD) card reader. The MOST Delta operates with 12-tooth T5 belts at 53.33 steps/mm for a Z precision of about 19 microns. The resolution of the printer in an X-Y plane is a function of distance from apexes, so it changes with distance from the center of the build platform [56].

The camera is based on 1/2.9 inch (6.23 mm in diagonal) Sony IMX322 CMOS Image Sensor [57]. This sensor consists of 2.24M square 2.8x2.8 µm pixels, 2000 pixels per horizontal line and 1121 pixels per vertical line. IMX322 has a Bayer RGBG color filter pattern (50% green, 25% red, and 25% blue) with 0.46÷0.61 red-to-green and 0.34÷0.49 blue-to-green sensitivity ratios. In operating mode, the camera captures 1280x720 pixel frames at a frequency of 30 Hz. The camera was calibrated on a widely used asymmetric circular grid pattern [58] since a chessboard pattern can be prone to uncertainties at chess blocks corner locations. The circle grid pattern can provide more accuracy and stability since the calibration technique is based on the detection of the center of gravity of each circle [59, 60].

**Table 1.** 3-D printing parameters allowing failure correction

| | | | MECHANICAL PARAMETERS | | | | | | | | TEMP. | | SLICER PARAMETERS | | | | |
|---|---|---|---|---|---|---|---|---|---|---|---|---|---|---|---|---|---|
| | | | A | B | C | D | E | F | G | H | I | J | K | L | M | N | P |
| LAYER OF OCCURRENCE | FAILURE NUMBER | FAILURE TYPE | CAD DESIGN | AMBIENT ENVIRONMENT | NOZZLE DIAMETER | ASSEMBLY / GEOMETRY | EXTRUDER PARAMETERS | AXES MOTOR PARAMETERS | BELT SLIPPAGE | MATERIAL CHARACTERISTICS | NOZZLE TEMPERATURE | BED TEMPERATURE | LAYER HEIGHT | SUPPORT PARAMETERS | INFILL PARAMETERS | EXTRUSION RATE | PRINT SPEED |
| INITIAL | 1 | Lost dimensional accuracy | | X | | X | X | X | | | X | X | | | | X | X |
| INITIAL | 2 | Circles are not round | | | X | X | | X | | | | | | | | | |
| INITIAL | 3 | Bed leveling issue | | | | | | | | | | | X | | | | |
| MEDIUM | 4 | Blocked nozzle | | X | X | | X | | | | X | | | | | X | X |
| MEDIUM | 5 | Adhesion problem (warping) | | X | | X | | | | | X | X | X | X | X | X | |
| MEDIUM | 6 | Print is not sticking to the bed | | | | X | X | | | | X | X | | X | X | X | |
| MEDIUM | 7 | Print offset / bending | | X | X | X | X | X | X | X | X | X | | | | X | X |
| MEDIUM | 8 | Printer stringing and oozing | X | | X | | X | X | X | X | X | | | X | X | X | X |
| MEDIUM | 9 | Walls caving in | | X | | | | | | | X | X | | | X | X | X |
| MEDIUM | 10 | Weak or under-extruded infill | | | X | | X | | | | X | X | X | | | X | X |
| MEDIUM | 11 | Deformed infill | | X | X | | X | X | | X | X | | X | | X | X | X |
| MEDIUM | 12 | Blobs in filament | X | X | X | X | X | X | | X | X | | X | X | X | X | X |
| MEDIUM | 13 | Small features are not printing | | X | X | X | X | X | X | X | X | X | X | X | X | X | X |
| MEDIUM | 14 | Temperature variations | | X | | | | | | | X | X | | | | X | X |
| MEDIUM | 15 | Poor bridging | X | | X | X | X | X | X | X | X | | | X | X | X | X |
| MEDIUM | 16 | Burnt filament blobs | X | X | X | X | X | X | X | X | X | X | | X | X | X | X |
| MEDIUM | 17 | Support falls apart | X | | X | X | X | X | X | X | X | X | | X | | X | X |
| MEDIUM | 18 | Incomplete infill | X | X | X | X | X | X | X | X | X | X | X | X | X | X | X |
| MEDIUM | 19 | Cracks in tall objects | | X | | X | X | X | X | X | X | X | X | | | X | |
| MEDIUM | 20 | Under extrusion | | | X | X | X | X | | X | X | | | | | X | X |
| MEDIUM | 21 | Over extrusion | | | X | X | X | X | | X | X | | | | | X | X |
| MEDIUM | 22 | Overhangs | X | X | | X | X | X | X | X | X | X | | X | X | X | X |
| MEDIUM | 23 | Missing layers | | X | X | X | X | X | X | X | X | X | X | X | X | X | X |
| MEDIUM | 24 | Overheating | | X | | | | | | | X | X | | X | X | X | X |
| TOP | 25 | Poor surf. quality above supports | | | | X | X | X | | | X | | | X | X | X | X |
| TOP | 26 | Gaps between infill and shell | | | X | X | X | X | | | X | | | | X | X | X |

The printing area is under monocular surveillance (Figure 1), where the single camera provides a rectified top view and pseudo-side-view of the printed part (Figure 3).

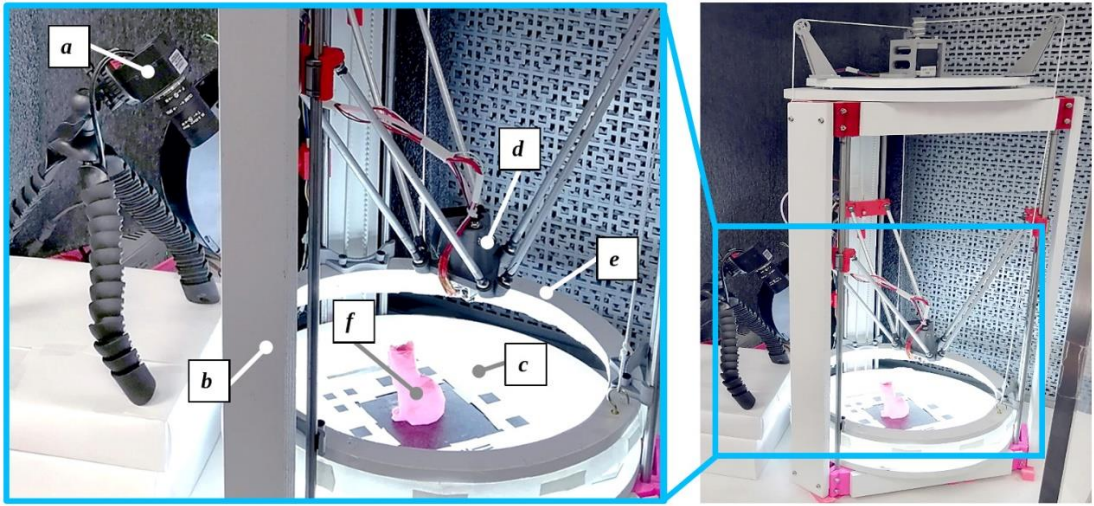

*Figure 1. Visual Servoing Platform: working area (left), printer assembly (right): a – camera; b – 3-D printer frame; c – visual marker plate on top of the printing bed; d – extruder; e – movable lighting frame; f – printed part.*

A developed visual marker plate located on top of the printing bed (Figure 1) enables the determination of the spatial position of the working area relative to the camera. The plate has a 88x88 mm printing area where seven contrast square markers (15x15 mm and 10x10 mm) build a reference frame for the camera. A computed camera pose in homogeneous coordinates allows the computation of a one-to-one relationship between the Euclidean world points $\mathbb{R}^n$ recorded in the G-Code or in the STL model and the planar image points captured by the camera by applying projective transformations (Figure 2). In computer vision problems, projective transformations are used as a convenient way of representing the real 3-D world by extending it to the three-dimensional projective space $\mathbb{P}^n$, where its points are homogeneous vectors [61, 62].

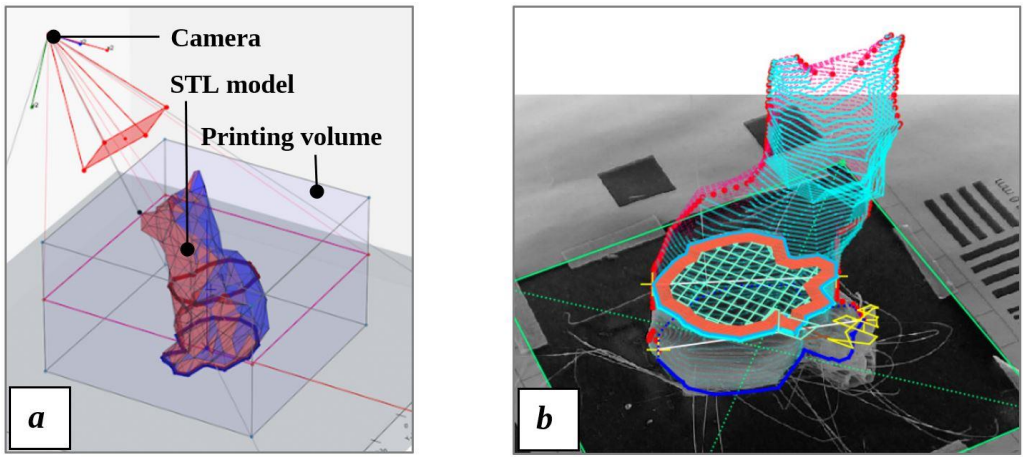

*Figure 2. Projective transformation of the G-Code and STL model applied to the source image frame: a – camera position relative to the STL model; b – G-Code trajectories projected on the source image frame. This and the following slides illustrate the printing analysis for a low polygonal fox model [63].*

The image pixel positions correspond to their three-dimensional spatial locations in accordance with the following equation (1), where the index *p* means "picture plane", and the index *w* means "world space":

$$\begin{bmatrix} x_p \\ y_p \\ 1 \end{bmatrix} = \begin{bmatrix} f_x & 0 & c_x \\ 0 & f_y & c_y \\ 0 & 0 & 1 \end{bmatrix} \cdot \begin{bmatrix} 1 & 0 & 0 & 0 \\ 0 & 1 & 0 & 0 \\ 0 & 0 & 1 & 0 \end{bmatrix} \cdot \begin{bmatrix} r_{11} & r_{12} & r_{13} & t_x \\ r_{21} & r_{22} & r_{23} & t_y \\ r_{31} & r_{32} & r_{33} & t_z \\ 0 & 0 & 0 & 1 \end{bmatrix} \cdot \begin{bmatrix} X_w \\ Y_w \\ Z_w \\ 1 \end{bmatrix}, \quad (1)$$

where $K = \begin{bmatrix} f_x & 0 & c_x \\ 0 & f_y & c_y \\ 0 & 0 & 1 \end{bmatrix}$ is the intrinsic camera parameters obtained during calibration, $f_x$ and $f_y$ are the focal lengths in image coordinates, $c_x$ and $c_y$ are the coordinates of the optical center in image coordinates (the principal point), $\mathbf{R} = \begin{bmatrix} r_{11} & r_{12} & r_{13} \\ r_{21} & r_{22} & r_{23} \\ r_{31} & r_{32} & r_{33} \end{bmatrix}$ is the rotation matrix, and $\mathbf{t} = [t_x \quad t_y \quad t_z]^T$ is the translation vector.

Applying projective transformations to rectified frames of the observed printing area, it is possible to obtain a virtual top-view as if the camera is mounted parallel to the normal vector of the printing bed [44] and a pseudo-side-view as if the camera is mounted perpendicular to the normal vector of the printing bed (Figure 3). Observing the printed layer through the camera lens, a slice of the material is viewed as a set of pixels, or a tuple of numbers, characterizing the local areas of the layer. Therefore, analyzing a two-dimensional image provides an understanding of the nature of the texture of the object.

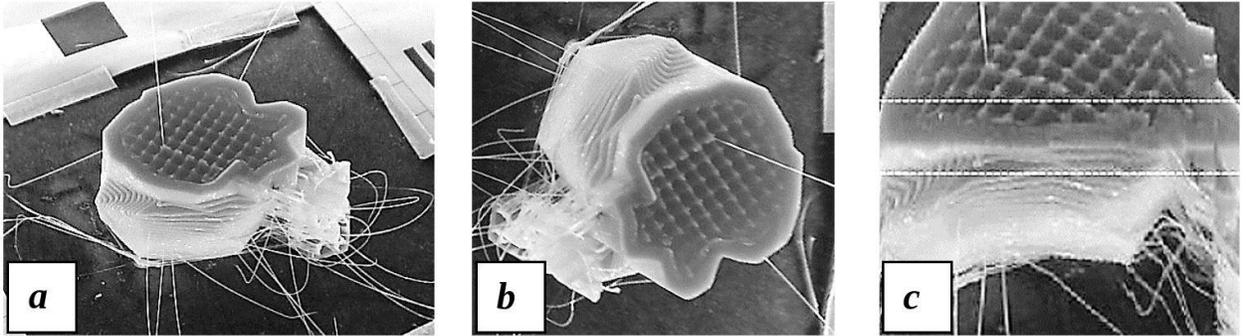

*Figure 3. Frames obtained from the monocular vision control system: a – rectified source image frame; b – unwrapped virtual top-view; c – unwrapped pseudo-side-view.*

After each layer, based on the 3-D printed layer height, an analytical projective plane in the image shifts accordingly with the layer number, so the rectified image frame remains orthogonal to the optical axis of the virtual top-camera. Thus, by utilizing a rich group of image processing techniques, it becomes possible to segment meaningful contour and texture regions based on images and known parameters of the STL model and the G-Code of the printing object. At the end of the printing process, a layered set of images provides additional data for volumetric analysis of the printed object in the future (Figure 4).

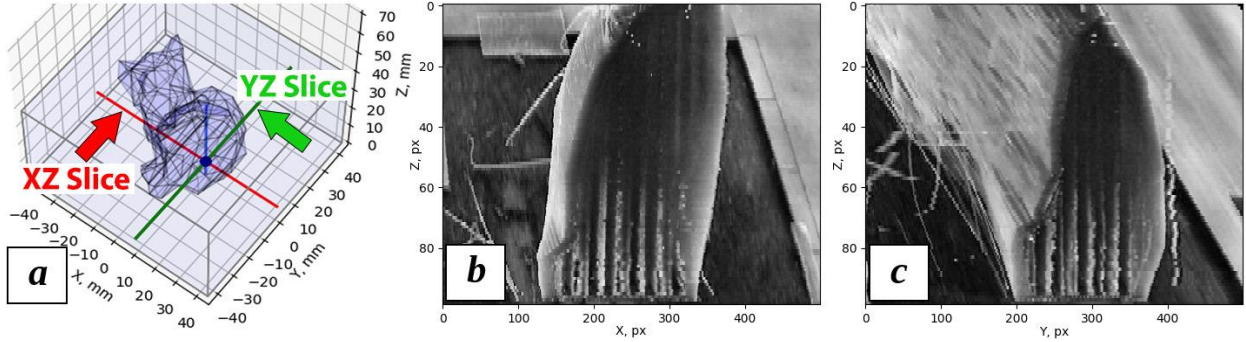

*Figure 4. Volumetric analysis of the printed part: a – STL model with the orthogonal scan planes; b – vertical XZ slice; c – vertical YZ slice.*

A movable circle-shaped lighting frame is installed above the bed surface. The motor and mechanical system for tensioning the cables are mounted on top of the printer. The motor is connected to the stepper motor driver in the RAMPS printer control system and drives a lighting frame, which rises with each newly printed layer to a distance equal to this layer height, which ensures constant and uniform illumination of the printed part. The lighting frame, in turn, is a circular set of 56 2.8×3.5 mm surface-mount device (SMD) light emitting diodes (LEDs) with a glowing temperature of 6000 K, a total luminous flux of 280 lumens, a power of 18 watts, and a supply voltage of 12 volts [64].

The software developed in Python-language environment parses the source G-Code, dividing it into layers and segmenting the extruder paths into categories such as a skirt, infill, outer and inner walls, support, etc. (Figure 5).

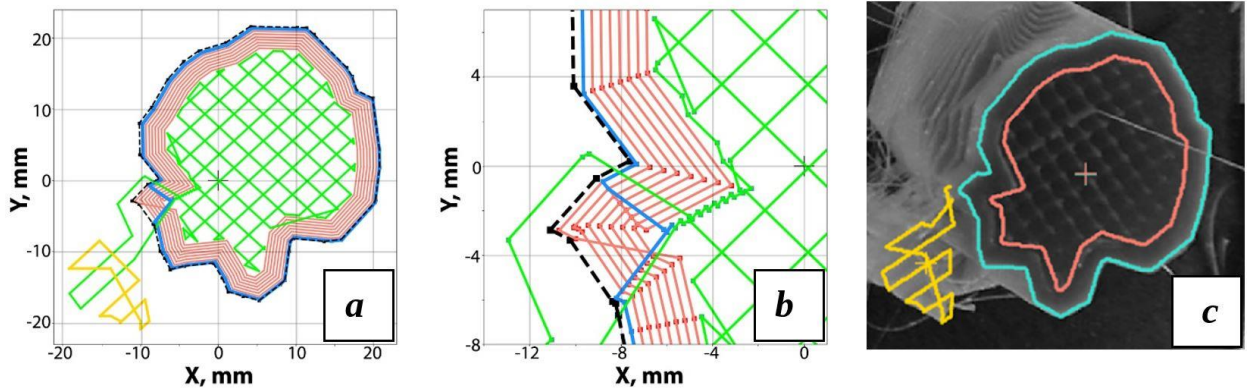

*Figure 5. Layer-wise G-Code segmentation: a – an STL layer outline (black dashed line) overlay with the segmented layer of the source G-Code with such trajectory categories as outer (blue) and inner walls (red), infill (green), and support (yellow); b – possible inconsistency of STL and G-code; c – unwrapped virtual top-view overlay with the outer wall (blue), infill (green), and support (yellow) extruder trajectories obtained from the G-Code.*

The categories of the G-Code paths depend on the software algorithm used to slice the STL model. The G-Code trajectories for any layer may not match exactly with the layer outline obtained from the STL model, and the magnitude of the mismatch can reach several millimeters (Figure 5, b). The STL model thus becomes unreliable and the G-Code coordinates are a reference source of positioning. The developed program synchronized with the printer uses RAMPS 1.4 3-D printer control system and the open-source firmware Marlin [65] as an intermediate driver. All source code is available: https://osf.io/8ubgn/ under the open source license GPLv3.

Using the reference layer-by-layer paths of the extruder obtained by the G-Code analysis, in addition to the virtual top view, it is also possible to generate a pseudo-side-view. This approach does not allow for the creation of a full scan of the entire side surface of the model, however, this provides an opportunity to assess the divergence in the vertical size of the part with the reference height obtained from the G-Code. The use of one camera instead of two (the main and a secondary camera for a side view) reduces the computational load and eliminates the need to synchronize the processing of two images (as well as reducing the hardware costs).

The temperature parameters, the coordinates of the trajectories and the traveling speed of the extruder, the feed rate of the material, as well as thickness of the printed lines and the height of the layer are stored in the program memory for each layer. Print commands, pauses, image analysis, and decision-making algorithms are carried out by the developed software, giving it full control over the state of the printing process. Therefore, in case of critical deviations, printing can be suspended, and if it is possible to repair the part, a sequence of corrective G-Code commands will be launched.

**3. Algorithm development**

The image processing pipeline could be divided into three branches and presented in Figure 6. Each branch of the pipeline will be described further in this paper.

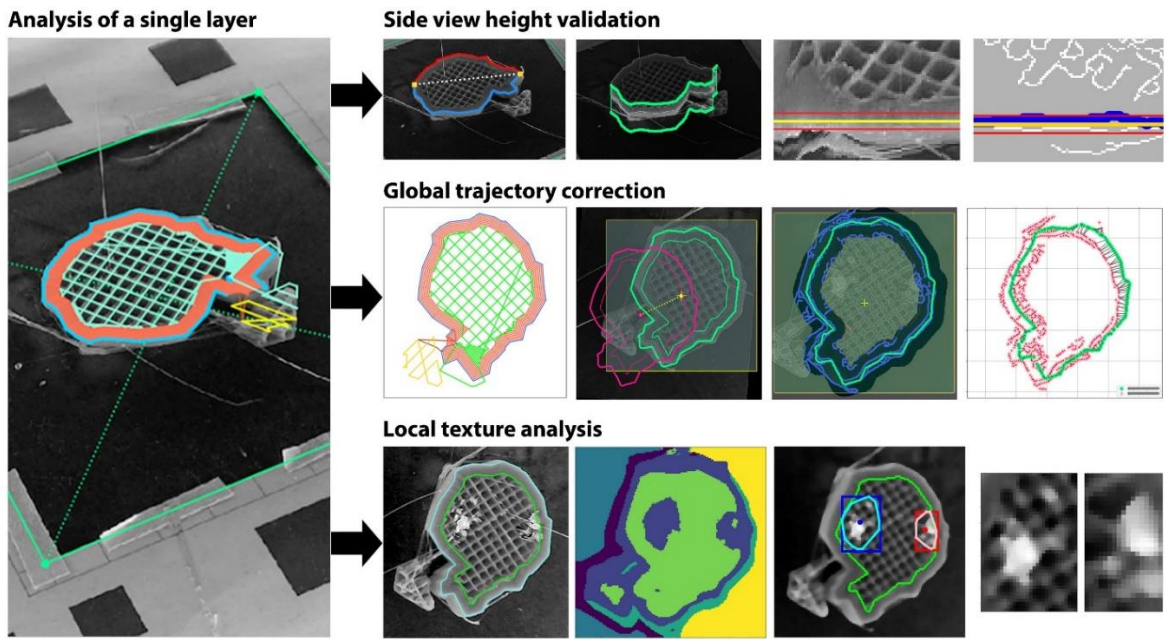

*Figure 6. Image processing pipeline*

Starting with the side-view height validation, the algorithm analyzes the virtual top view for global trajectory matching and local texture examination. This allows taking into account both global and local parameters of printing processes.

The proposed algorithm (Figure 7) for detecting printing failures assumes the presence of one camera located at an angle to the working surface of the 3-D printer. An angled camera allows us to observe both the active printable layer and part of the printing model from the side. Thus, one source frame can be divided into a virtual top view from above and a pseudo-view from the side.

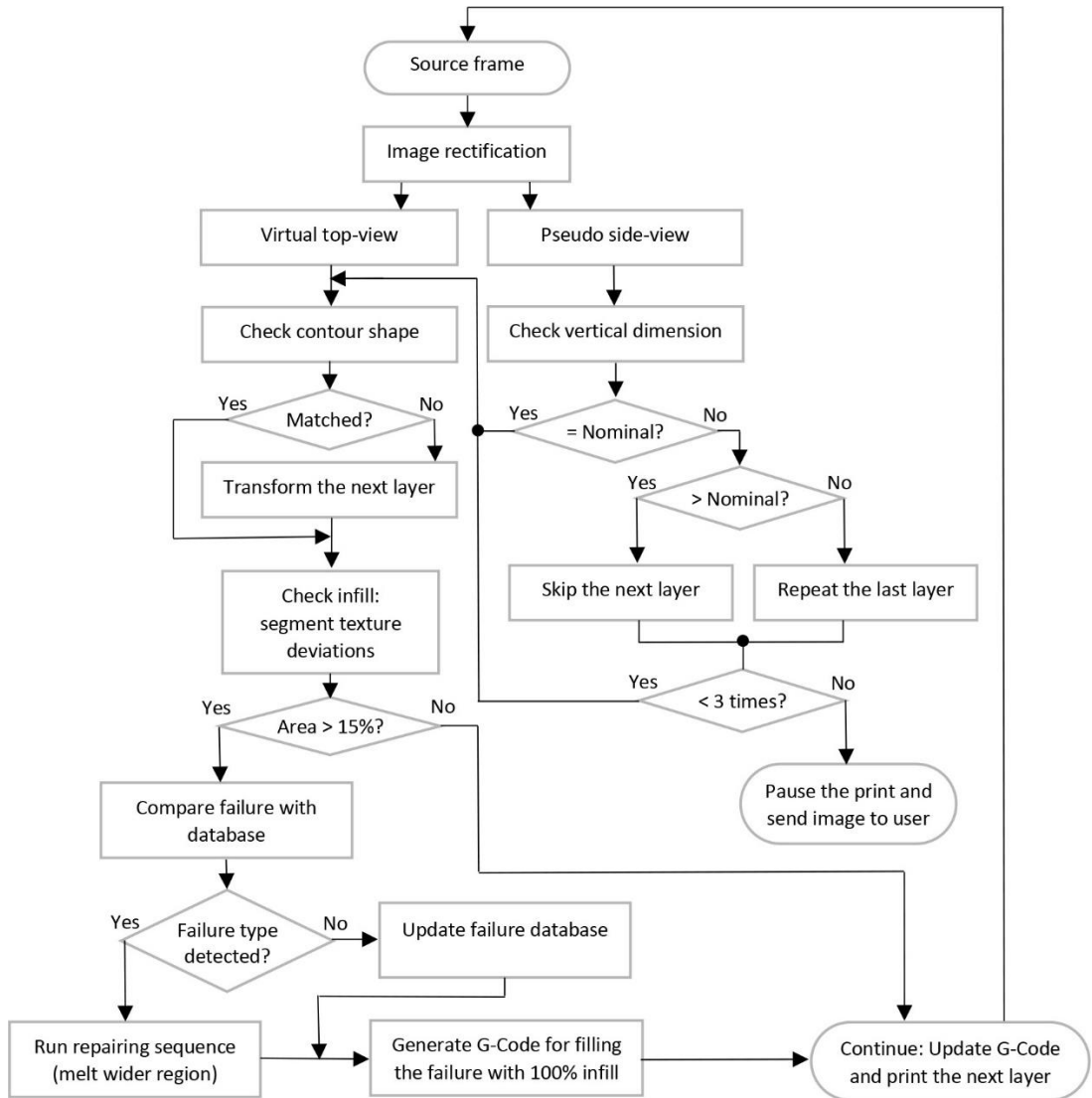

*Figure 7. 3-D printing control algorithm.*

Such criteria as bed leveling, dimensionality lost, and non-circularity are dependent on a specific printer model and are manually calibrated by the user at the time of the first run. It is possible to create calibration tables to determine the correction factors for G-Code trajectories. However, at this stage, the above parameters are checked only for compliance/non-compliance with the specified values. In case of non-compliance in bed leveling, dimensionality, and circularity, printing is suspended. This method does not eliminate these errors during the printing process, but it saves time and material.

**3.1. Side view height validation**
Knowing the camera position and G-Code trajectories for a given layer, it is possible to analyze the visibility of the side area of the printed part by solving the system of equations (2) which provides slope and shift coefficients for a linear visibility delimiter.

$$\begin{cases} y_p^{(1)} = m \cdot x_p^{(min)} + b \\ y_p^{(2)} = m \cdot x_p^{(max)} + b \end{cases}, \qquad (2)$$

where $m$ and $b$ – are the coefficients of the linear visibility delimiter, $x_p^{(min)}$ and $x_p^{(max)}$ obtained from the extreme contour points of the projection of the G-Code outline on the picture plane, $y_p^{(1)}$ and $y_p^{(2)}$ – are the y-coordinates of the corresponding extreme points on the picture plane.

The pseudo-side-view allows monitoring the height of the printed part and detect critical failures such as "blocked nozzle", "lack of material", "major deformations", etc.

**3.2. Global trajectory correction**
After checking the vertical size, a virtual top view is used for the subsequent two-stage analysis of the external contour and infill of the printed layer. Having data on the corresponding extruder trajectories from the G-Code and the resulting contour, it is possible to determine if there is a mismatch between the real outline and the reference borders using the Multi-Template Matching (MTM) [66] and the Iterative Closest Point (ICP) [68, 69] algorithms. MTM allows to track significant horizontal and vertical displacements of the printed part based on the binary layer template obtained from the G-Code trajectories, and the ICP algorithm determines fine rotation and translation within the small deviation range. As a result, we obtain a transformation matrix, which multiplied by the spatial coordinates of the extruder trajectories, eliminates small linear shifts and scale distortions of the printed layer.

The MTM method [66] computes a correlation map between the reference layer outline and binary edge image of the virtual top view based on the *"match Template"* OpenCV function [67] and predicts the template (G-Code outline) position within the image. Since the algorithm performs the search by sliding the template over the image, it detects the object with a similar orientation as the template and may not be sensitive to rotations.

The ICP algorithm [68, 69] aimed at finding the transformation matrix between two point clouds by minimizing the squared error (3) between the corresponding surfaces using the gradient descent method. The iterative algorithm converges when the starting positions are close to each other.

Given two corresponding point sets *{m₁, m₂, ..., mₙ}* and *{p₁, p₂, ..., pₙ}* we can find translation *t*, rotation **R**, and scaling *s* that minimize the sum of the squared error:

$$E(\mathbf{R}, \mathbf{t}, s) = \frac{1}{N_p} \sum_{i=1}^{N_p} \|m_i - s\mathbf{R}(p_i) - \mathbf{t}\|^2, \tag{3}$$

where **R**, **t**, and *s* – are rotation, translation, and scaling respectively. Since the scaling operation can also translate an object, the center of the G-Code layer projection should be placed at the origin.

Based on rotation, scaling and translation obtained from the ICP algorithm and assuming that the z-level was found during the vertical size check, the G-Code trajectories for the next *(k+1)*ᵗʰ layer $\left[G_x^{(k+1)\prime} \quad G_y^{(k+1)\prime}\right]^T$ will be transformed from the initial trajectories of the next layer $\left[G_x^{(k+1)} \quad G_y^{(k+1)}\right]^T$ in accordance with the following equation (4):

$$\begin{bmatrix} G_x^{(k+1)\prime} \\ G_y^{(k+1)\prime} \end{bmatrix} = \begin{bmatrix} s_x & 0 \\ 0 & s_y \end{bmatrix} \cdot \begin{bmatrix} \cos\theta & -\sin\theta & t_x \\ \sin\theta & \cos\theta & t_y \end{bmatrix} \cdot \begin{bmatrix} G_x^{(k+1)} \\ G_y^{(k+1)} \\ 1 \end{bmatrix} \tag{4}$$

### 3.3. Local texture analysis

After analyzing the contour, a check is made to the layer filled with the material. The purpose of this step is to identify irregular sections of the texture within the layer infill. At this stage, it is assumed that the vertical dimension of the part corresponds to the specified one, and the correct location of the real boundaries of the part is determined. Thus, only irregularities of the texture in the region bounded by the outer shell of the layer are considered.

The textural features based on local probabilistic similarity measures are simple, have a low computational load, and could serve well for a small number of specific cases, but may not be efficient for a wide range of real-world problems [71-73]. Because of the complex surface topology, textural variations may not be explicitly expressed as a histogram comparison [74].

In this work, the texton-based approach to texture segmentation was implemented, since the given method has repeatedly demonstrated its effectiveness and scalability [74-78]. The texton-based segmentation utilizes Leung-Malik (LM) filters [74, 79] (Figure 8) which work as visual cortex cells and allow the segmentation of an image into channels of coherent brightness and texture in a natural manner, where the texture originates from the spatial variation of surface normal and reflectance. The LM filter bank consists of 48 filter kernels, which is a mix of 36 elongated filters, 8 difference of Gaussians, and 4 low-pass Gaussian filters [74].

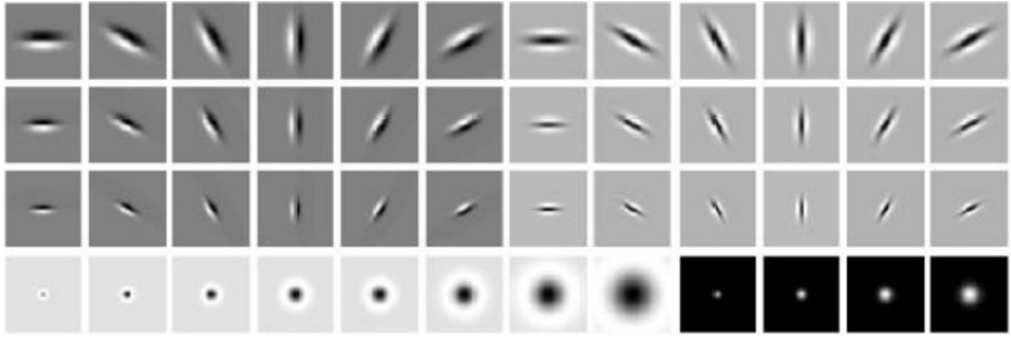

*Figure 8. The Leung-Malik orientation and spatial-frequency selective filter bank [74, 79]*

After the convolution of an image with the filter kernels, each pixel is transformed into a multidimensional vector of filter responses. After clustering, these vectors form a set of texton channels, or appearance vectors, that define the image partitioning.

In both plastic and metal additive manufacturing the lighting effects and mutual reflectance create a non-Lambertian environment, where similar surfaces may look significantly different under varying viewing angles [74,78], which narrows the set of possible image processing techniques. According to [75, 81-83], filter responses encode appearance information over a broad-scale range and can serve as a preprocessing method that can be combined with dense descriptors for efficient texture classification with varying illumination conditions.

Without prior texture information, a suitable approach for clustering the obtained filter responses, and, therefore, for texture segmentation, is an unsupervised machine learning method, which is a widely used technique in object recognition. Unsupervised machine learning is used to draw derivations from data consisting of input information with no labeled responses.

Clustering, being the most common unsupervised learning method, aimed at finding hidden patterns in the source data set and grouping them accordingly to their salient features and the number of clusters *k* specified by the user. The majority of previous works in texture segmentation are based on k-means clustering [75], but the non-probabilistic nature, hard cluster boundaries and lack of flexibility in cluster shape in k-means clustering leads to practical challenges and may not perform well in real-world applications. The Gaussian Mixture Model (GMM) clustering, implemented in this work, may give better performance.

The GMM clustering is based on filter responses, and attempts to partition an unlabeled input layer texture as a mixture of multidimensional Gaussian probability distributions in regions that share common characteristics. The Gaussian mixture model for *k* unknown clusters can be written as a superposition of Gaussians [84]:

$$f_{GMM}(x) = \sum_{j=1}^{k} w_j f_{\mathcal{N}(\mu_j, \Sigma_j)}(x) \text{ , where } \sum_{j=1}^{k} w_j = 1, \tag{5}$$

which is a combination of weighted $w_j$ normal probability density functions $f_{\mathcal{N}(\mu_j, \Sigma_j)}$ with mean vector $\mu_j$ and covariance matrix $\Sigma_j$.

The GMM-based clustering determines the maximum likelihood for Gaussian models with latent variables by utilizing the Expectation-Maximization algorithm [85, 86], which iteratively estimates the means, covariances, and weighting coefficients in a way that each cluster is associated with a smooth Gaussian model. However, despite the effectiveness of the described method, the exact number of clusters $k$ should be specified in advance, which is a critical decision and determines the success of texture segmentation.

Using the infill mask obtained from the G-code paths, GMM segmentation and failure analysis are performed only within the layer infill texture region. Since the result of segmentation is not completely predictable and one anomalous region can consist of several segments, the Agglomerative Hierarchical Clustering (AHC) [87] is launched after the GMM partitioning. The AHC, being an unsupervised technique, recursively merges pairs of individual clusters based on their location within the segmented area, which provides a consistent failure map for the analyzed infill region.

### 3.4. Targeted failures and corrective actions

Using the above-mentioned techniques, it is possible to approach a considerable number of the failures listed in Table 1. The main target failures along with the proposed corrective actions introduced in Table 2. These printing errors can be detected and/or eliminated by using the appropriate G-Code commands without adjusting the mechanical parameters of the printer and slicing modes.

**Table 2.** Target failures and corrective actions

|   | Failure type | Detection strategy | Printer action |
|---|---|---|---|
| 1 | Out of filament | Vertical level + MTM & ICP algorithms | Pause / Report |
| 2 | Blocked nozzle | Vertical level + MTM & ICP algorithms | Increase nozzle temperature; Repeat the previous layer a finite number of times |
| 3 | Missing layer | Vertical level + MTM & ICP algorithms | Repeat the previous layer a finite number of times |
| 4 | Lost dimensional accuracy | MTM & ICP algorithms | Update G-Code coordinates |
| 5 | Bed leveling issue | Texture segmentation | Pause / Report; Manual level recalibration |
| 6 | Adhesion problem (warping) | Tracking vertical level of the initial layer | Increase bed temperature; Pause / Report in case of critical vertical deviation |
| 7 | Print is not sticking to the bed | Vertical level + MTM & ICP algorithms | Increase bed temperature; Pause / Report |
| 8 | Print offset / bending | Vertical level + MTM & ICP algorithms | Update G-Code coordinates |
| 9 | Weak or under-extruded infill | Texture segmentation | Increase nozzle temperature and feed rate |

**Table 2, continued**

|    | Failure type | Detection strategy | Printer action |
|----|---|---|---|
| 10 | Deformed infill | Texture segmentation | Change nozzle temperature and feed rate |
| 11 | Burnt filament blobs | Texture segmentation | Ironing |
| 12 | Incomplete infill | Texture segmentation | Patch replacement procedure |
| 13 | Poor surface quality above supports | Texture segmentation | Change feed rate |
| 14 | Gaps between infill and shell | Texture segmentation | Change feed rate |

A simplified pseudo-code in accordance with the basic algorithm (Figure 7) for a layer processing is shown in Figure 9.

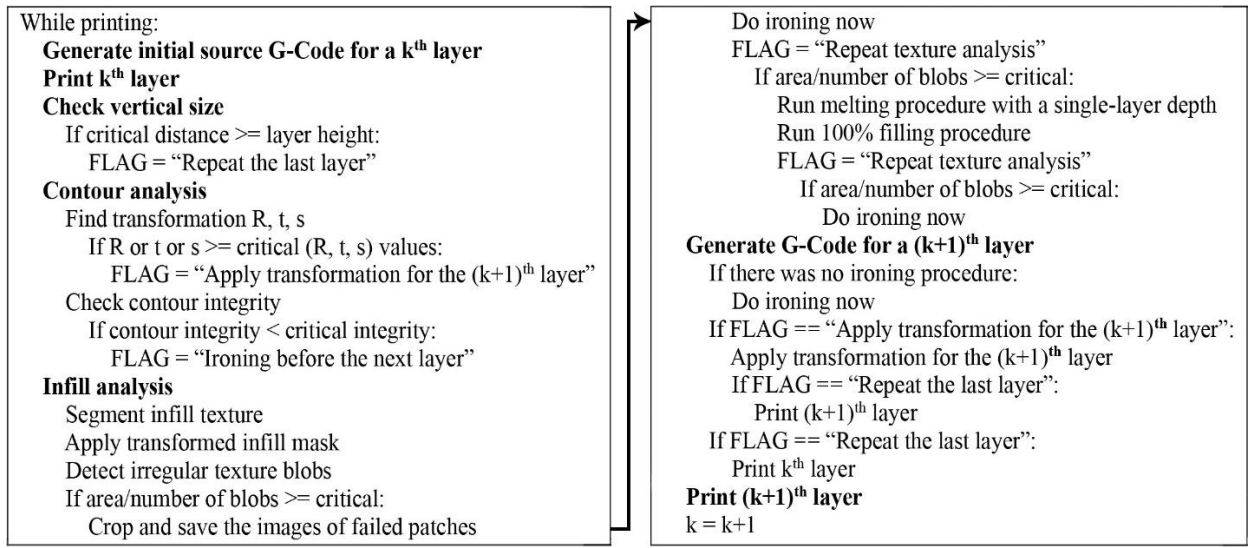

*Figure 9. Example of failure correction*

## 4. Experimental results

The algorithm was tested during regular printing without failures of the 42x51x70 mm low-polygonal fox model [63] with the following printing parameters: 1.75 mm PLA, 0.4 mm layer height, 0.4 mm line width, 30% grid infill, and 3.2 mm wall thickness. The entire model consists of 175 layers, but the tests were carried out for the first 96 layers since part of the model was located outside of the visible area.

Bed leveling and dimensionality checks were calibrated in advance before printing. The visual analysis of the vertical level, deformation of the outer shell, and infill texture was performed for each printing layer on a 2.5 GHz processor with 8 GB of RAM. These experiments for the case of a normal printing mode without deviations allow determining the accuracy and tolerance of the adaptive algorithm.

## 4.1. Height validation results

Starting with the vertical level validation (Figure 6), the algorithm analyzes the virtual top view for global trajectory matching and local texture examination. This allows taking into account both global and local parameters of printing processes. Based on the G-Code reference contour and the system of equations (2), the calculated visual separator is used to generate a pseudo-side-view projection where the curved part of the model is shown as a straight line (Figure 10).

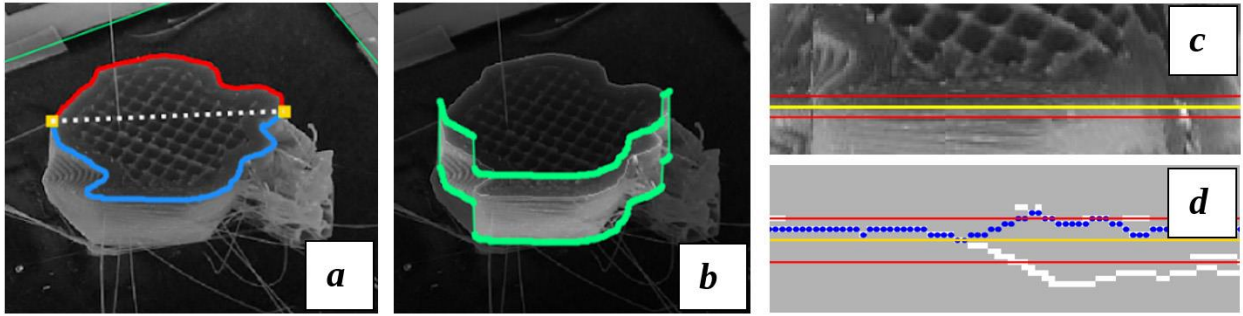

*Figure 10. Pseudo-side-view generation:* a – computed linear visibility delimiter (white dashed line), the edge for the visible side region (blue) and invisible side region (red) of the printed part; b – designated visible area for unwrapping; c – unwrapped region with the reference vertical level (yellow) and the maximum double-layer errors in both directions (red); d – detected vertical edge error (blue) with the reference layer height (yellow) and maximum double-layer errors in both directions (red).

Due to uniform all-round lighting, the contrast between adjacent faces of the part is enough to find a clear edge. Thus, comparing the pixel-wise deviation of the detected edge from the reference contour becomes possible to obtain the distribution and total vertical level error for each layer (Figure 11).

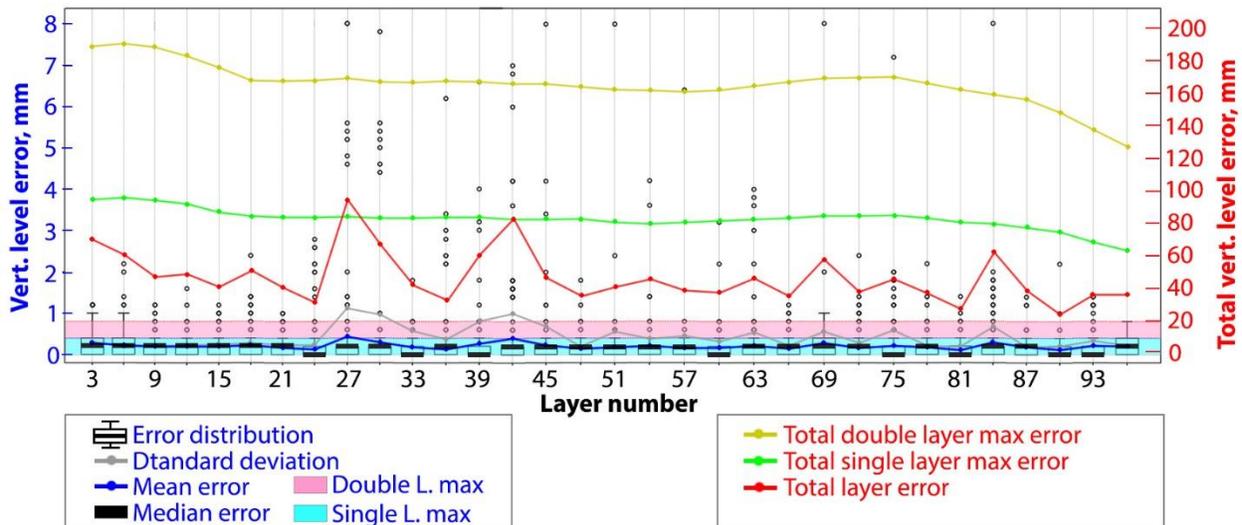

*Figure 11. Results of vertical level validation*

As can be seen from the experimental results, in the normal printing mode, both the average and total error values do not exceed the maximum deviation equal to the height of the two layers. For each individual layer, however, the vertical level error may exceed the maximum error value equal to the value of one layer. The discriminative power of the failure detection depends on the resolution of the camera, the distance to the print area, the size of the part, and can be taken into account in the algorithm in such a way that a one-time deviation exceeding the height of one layer will be ignored, while multiple consecutive excesses of the level of one layer will be taken as a true error.

### 4.2. Results of the global trajectory analysis

The global outline analysis uses a virtual top view and a combination of MTM and ICP algorithms. At the first stage, the MTM algorithm detects significant horizontal and vertical shifts (Figure 12), after which the ICP algorithm provides more accurate information about the rotation and displacement in a small range within the area detected by the MTM algorithm (Figure 13).

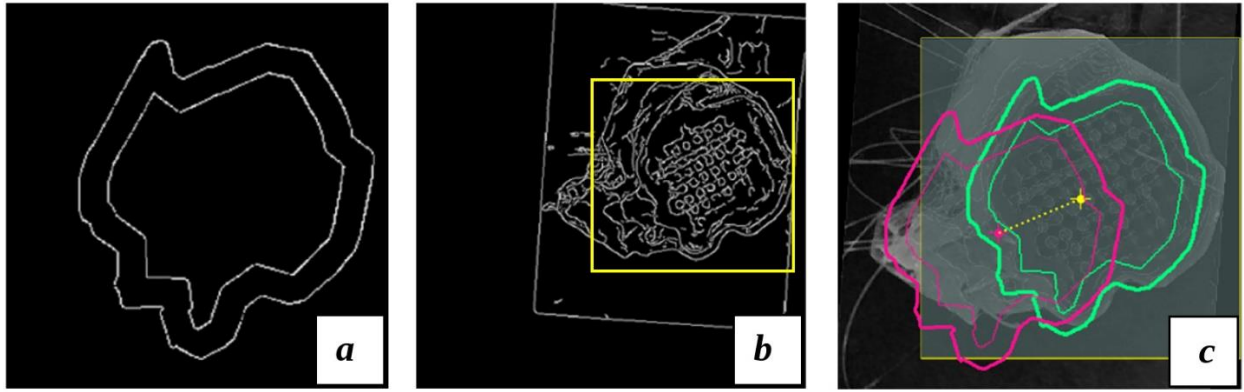

*Figure 12. Global displacement detection based on MTM algorithm:* a – *contour-based binary template; b – printed part shifted due to failure; c – computed shift distance and direction.*

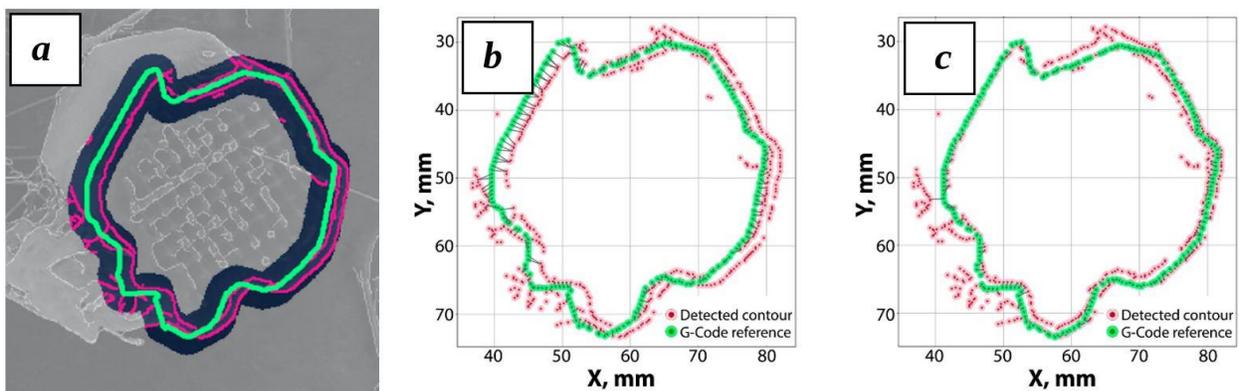

*Figure 13. Global G-Code trajectory matching for a single layer based on ICP algorithm:* a – *reference outline (green) mismatched with the detected contour points (red); b – initial ICP iteration; c – final ICP iteration.*

To ensure the reliability of the ICP algorithm, a restrictive mask based on the STL layer outline is used that limits the number of detected edge points used in the calculations of displacement and rotation relative to the G-Code reference contour (Figure 13, a).

Figure 14 shows the results of the ICP algorithm. It was revealed during the experiments that the restrictive mask with a width of 30 pixels (which corresponds to 5.7 mm for the given setup, or about 10% of the horizontal size of the printed part) allows to obtain a stable result and to identify the maximum displacement and rotation of 8 mm and 10 degrees, respectively.

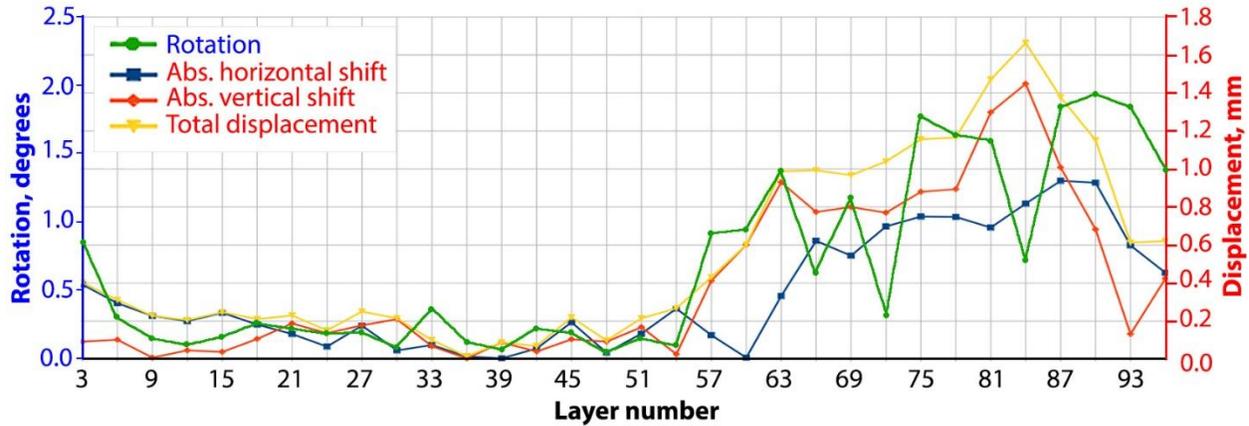

*Figure 14. Results of the global trajectory matching analysis*

In the nominal printing mode, rotation errors of up to 2 degrees and displacement errors of up to 1.7 mm were observed.

### 4.3. Results of the local texture analysis

According to [88], the dimensions of the filter should correlate with the size of prevailing image structures to reduce the noise effect during image processing. Varma and Zisserman in [89, 90] also presented a detailed analysis of the effect of filter size on classification efficiency for a number of filter banks and filter sizes.

Taking into account the facts that to accelerate the clustering operation, the original image can be reduced in size, and that a larger filter can better suppress visual noise, it was experimentally determined that the filter dimension of 1/3 of the input image allows to effectively segment the layer texture with high speed. Thus, the dimensions of the input image and the Leung-Malik filters are 150x150 pixels and 49x49 pixels, respectively.

In addition to the filter size, the number of expected texture clusters is also an equally important parameter. During the experiments was found that six clusters (six expected textures in the input image) can provide an effective ratio of speed and quality of segmentation.

Figure 15 shows the result of the GMM segmentation of the test image with 18 various texture samples, where the size of each texture patch is equal to the size of the filter and is 49x49 pixels.

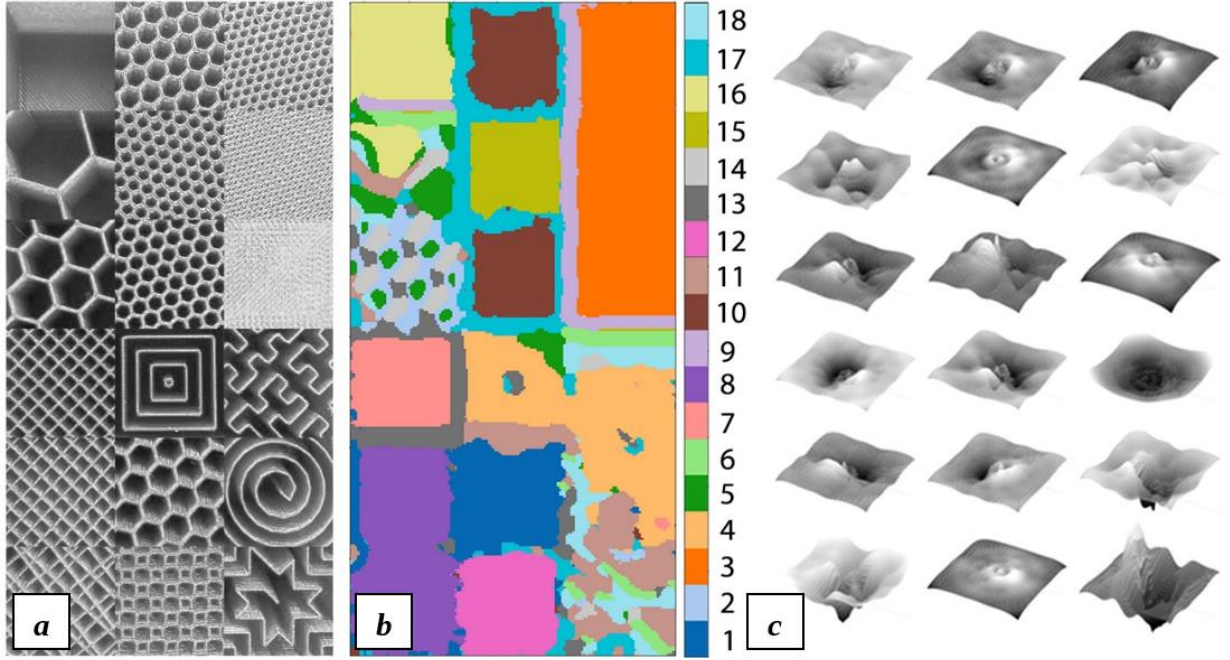

*Figure 15. Partitioning an image into texture channels (textons): a – test image with 18 various texture samples (modified image from https://all3dp.com/2/infill-3d-printing-what-it-means-and-how-to-use-it); b – segmented textures; c – obtained texture channels.*

The textons, appearance vectors, or the corresponding filter response vectors, capture characteristic shapes of different materials and features at various viewing angles and lighting conditions. In Figure 15 the 18 clusters correspond to the prevalent features in the image [74], where each of the cluster centers visualized by a pseudo-inverse which codes geometric features such as grooves, bumps, ridges, and hollows into an image icon [91].

After segmentation of the input image, labeling of the textures regions inside the infill mask occurs (Figure 16). Texture regions other than the main infill texture are considered abnormal and, subject to a certain shape and size, are taken into account in further analysis. If the total area of the considered anomalous regions exceeds a critical value of 15% of the total area of the layer infill region, the layer is considered defective, and further unsupervised agglomerative hierarchical clustering occurs.

Since after a GMM segmentation a defective part of a texture can consist of several segments, it is necessary to determine their belonging to one or another defective group. For the AHC algorithm, the clustering parameters were experimentally selected based on the centroid locations of the anomalous regions and the distance between them. The algorithm assumes one or two defective infill sections and determines whether the segmented anomalies belong to one or the other section.

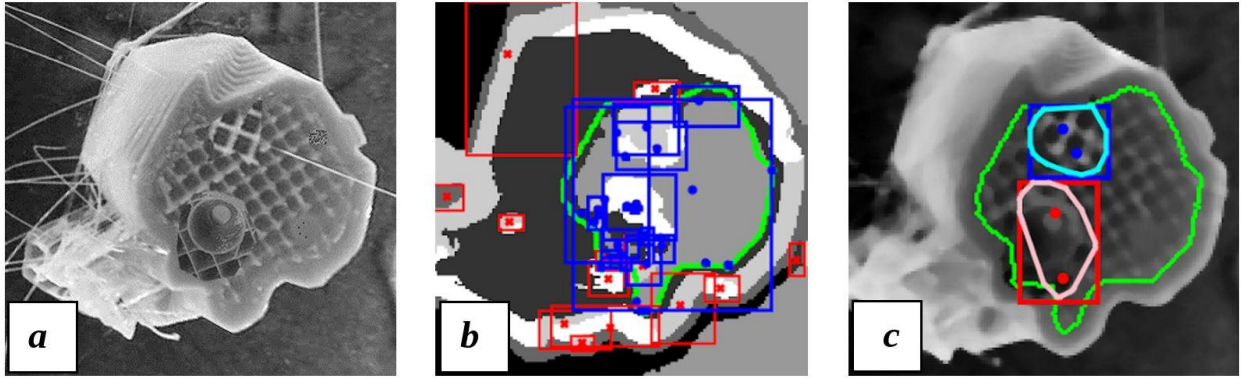

*Figure 16. Defective layer segmentation results: a – source virtual top view; b – segmented image with the with infill mask (green), infill textures (blue regions), outside textures (red regions); c – segmented failures (red and blue regions) inside the infill area (green).*

Figure 17 shows the results of the infill texture analysis for the specified printed layers. During the experiments, the entire total area of anomalous regions was considered without taking into account their shapes.

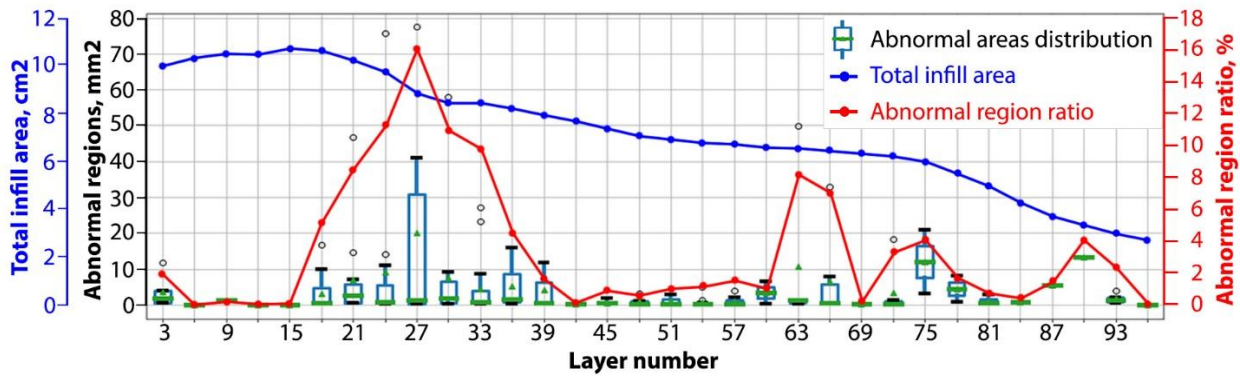

*Figure 17. Results of the texture analysis during the regular printing*

As can be seen from the figure, the total area of the anomalous regions can reach 10 or more percent of the infill region. This fact is due to the ingress of the outer wall texture inside the infill mask, which is a false alarm and can be eliminated by analyzing its shape.

### 4.4. Runtime analysis
The total analysis time can be divided into the following components (Figure 18):
1. side view height validation,
2. global outline correction,
3. local texture analysis.

The computational complexity of the side view height validation is $O(n)$ and depends on image resolution, so the runtime remains practically unchanged throughout the print cycle.

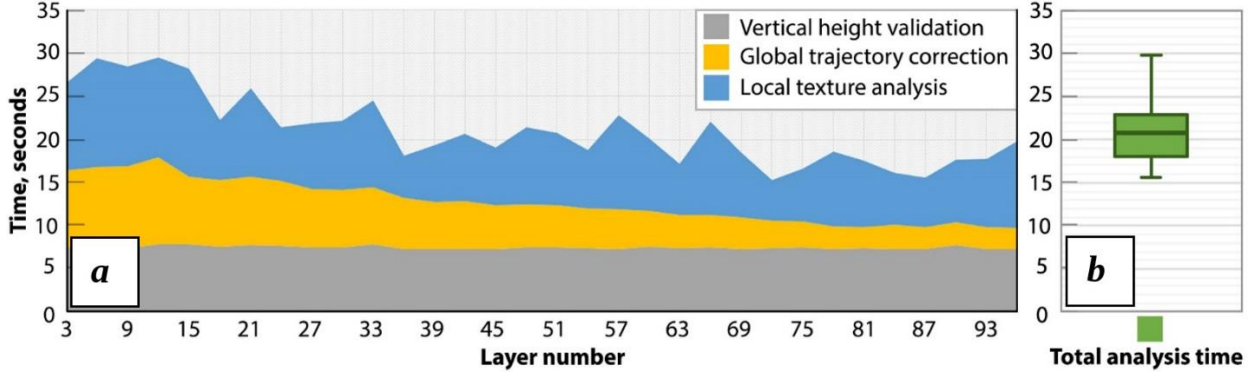

*Figure 18. Runtime distribution:* *a – time decomposition of layer-wise visual analysis;*
*b – distribution of the total analysis time for all layers.*

The global texture correction step consists of MTM and ICP algorithms, the computational complexity of which is $O(n^2 log^2 n)$ and $O(n^2)$, respectively. Thus, a significant reduction in the analysis time is observed due to a decrease in the geometrical dimensions of the part when approaching the upper printing layers, which, in turn, leads to a reduction in the number of data points involved in the contour transformation computations.

The final stage, local texture analysis, includes GMM clustering based on the EM algorithm with the computational complexity of $O(n)$, and failure segmentation based on the AHC algorithm with $O(n^3)$ complexity. The AHC algorithm, however, does not introduce a significant time delay due to the small number of texture centroids that make up the hierarchy. Due to the sporadic initialization of the center clusters during the GMM segmentation, however, the execution time of the Expectation-Maximization algorithm may vary over a wide range.

Regular printing time (without using the visual analysis) of the model [63] is 2 hours and 14 minutes (8,040 seconds). The total analysis time varies between 15 and 30 seconds with an average of 21.4 seconds. Thus, introducing the visual analysis in the printing process of the given part increases the total time of the production process by an amount of the order of 50% (6):

$$\frac{L_t \cdot N}{P_r} \cdot 100\% = \frac{21.4 \cdot 175}{8040} \cdot 100\% = 46.6\% , \qquad (6)$$

where $L_t$ – is the average layer analysis time, $N$ – is the total number of layers, $P_r$ – is the regular printing time.

Applying the above analysis to selective layers reflecting pivotal changes in the geometry of the part or in the temperature conditions of printing can offset the time costs and bring the analytical cycle closer to the real-time regime.

**4.5. Failures database and future development**
Examples of segmentation of artificially created print defects presented in Figure 19. One of the proposed methods for eliminating defects is the operation of ironing, followed by repeated segmentation and texture analysis. If the defective area is preserved, this area must be melted and filled with 100% material. A protocol for this procedure is under development.

At this stage of texture analysis, in addition to direct segmentation, images of defective areas are saved in the database for subsequent labeling and classification. In the future, possibly, appropriate correction procedures will be developed individually for each type of defect. Thus, the machine vision algorithm developed in this study will form the foundation for a fully open source failure correction algorithm for all of additive manufacturing.

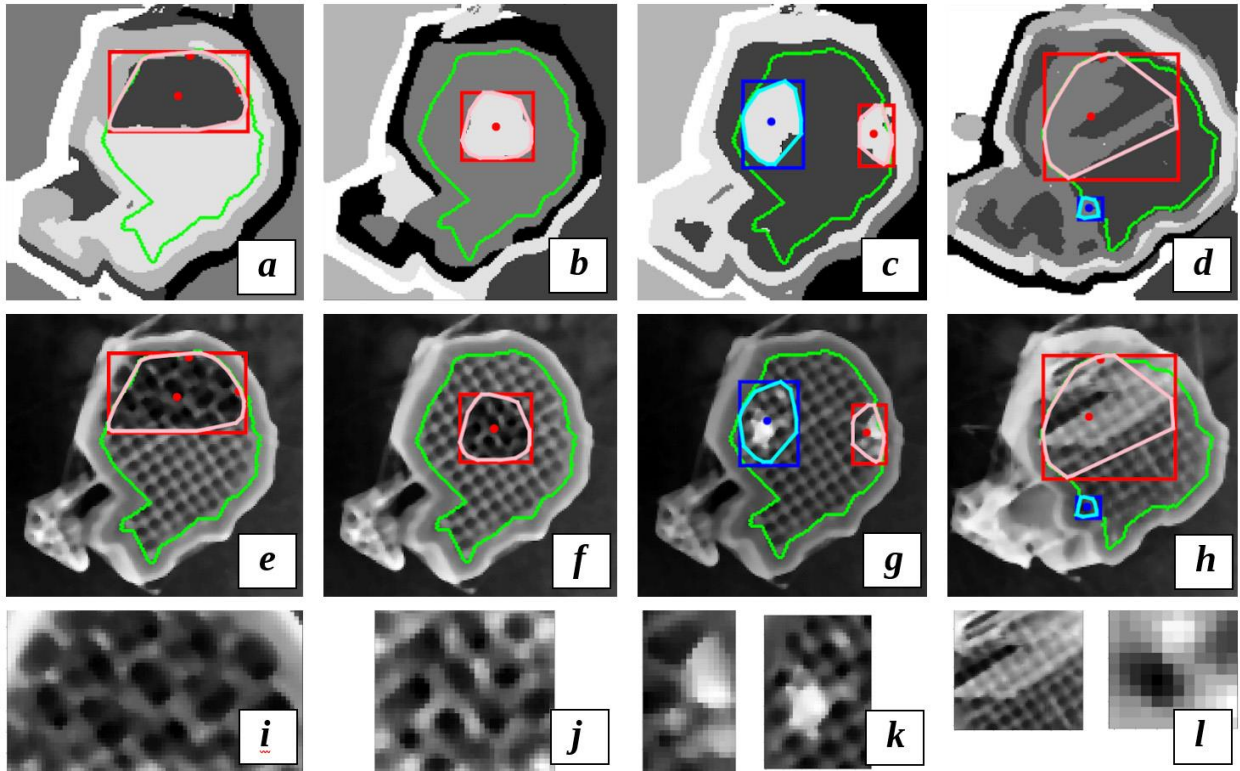

*Figure 19. Detected regions with abnormal texture: (a-d) – segmented textures; (e-h) – detected failures; (i-l) – cropped regions of interest with failures (not to scale).*

## 5. Conclusions

The development of an adaptive algorithm is a comprehensive and complex problem, because it is challenging to (1) uniquely visually determine the type of error, (2) establish a direct causal relationship between the type of error and the printing parameter involved, and (3) declare in advance what parameter value (scaling coefficients, feed rate, temperature, traveling speed, etc.) should be used to correct the failure.

The experiments above are based on the assumption that the mechanical parameters (stability of assembly, the presence of grease in moving parts, belt tension, the electrical voltage of stepper motor drivers, etc.) of the printer are configured and calibrated optimally. The experimental results obtained for the case of the nominal printing mode without deviations allow determining the accuracy and tolerance of the adaptive algorithm.

Thus, at this stage of the research, the presented work is more an intelligent printing suspension tool designed to save time and material rather than a full failure correction algorithm for printing enhancement. However, this work will allow users to systematize knowledge about failure mechanisms and will serve as a starting point for deep study in the future and a full failure correction system for open source additive manufacturing.

## 6. Acknowledgements

This work was supported by the Witte Endowment. The authors would like to acknowledge helpful discussions with Adam Pringle and Shane Oberloier. The authors also thank Eric Houck for assistance in developing a movable lighting frame for the 3-D printer.

[89]     M. Varma, A. Zisserman, 2003. Texture classification: are filter banks necessary?, In Proceedings of the IEEE Computer Society Conference on Computer Vision and Pattern Recognition, 2003, Madison, WI, USA, pp. II-691. https://doi.org/10.1109/CVPR.2003.1211534.
[90]     M. Varma, A. Zisserman, 2005. A Statistical Approach to Texture Classification from Single Images, Int J Comput Vision. 62, 61-81. https://doi.org/10.1023/B:VISI.0000046589.39864.ee.
[91]     S.-C. Zhu, C.-E. Guo, Y. Wang, Z. Xu, 2005. What are Textons? International Journal of Computer Vision 62(1/2), 121-143. https://doi.org/10.1007/s11263-005-4638-1.
29